\documentclass{article} 
\usepackage{colm2024_conference}
\usepackage{multirow}
\usepackage{microtype}
\usepackage{hyperref}
\usepackage{url}
\usepackage{booktabs}
\usepackage{graphicx}
\usepackage{wrapfig}
\usepackage{caption}
\usepackage{tabularx} 
\usepackage{booktabs}
\usepackage{url}
\title{\includegraphics[width=0.9cm]{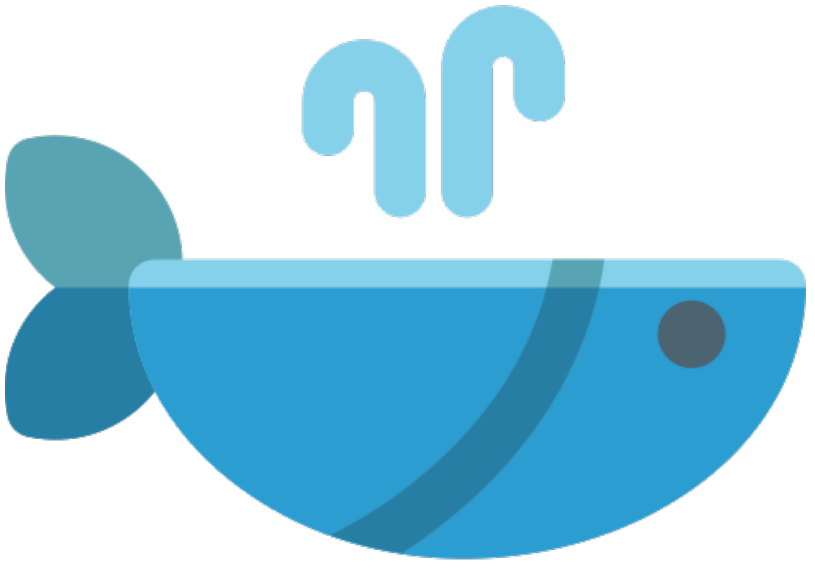} Kun: Answer Polishment for Chinese Self-Alignment \\with Instruction Back-Translation}

\colmfinalcopy
\author{$^1$\,$^5$Tianyu Zheng\footnotemark[1]\;,
    $^5$Shuyue Guo\footnotemark[1]\;\\[0.5mm]
    $^4$\,$^5$\textbf{Xingwei Qu},
    $^5$\textbf{Jiawei Guo},
    $^1$\,$^5$\textbf{Xinrun Du},
    $^1$\textbf{Qi Jia}\\[0.5mm]
    $^4$\,$^5$\textbf{Chenghua Lin},
    $^1$\,$^5$\textbf{Wenhao Huang}\footnotemark[2]\;, 
    $^3$\textbf{Jie Fu}\footnotemark[2]\;,
    $^1$\,$^2$\,$^5$\textbf{Ge Zhang}\footnotemark[1] \, \footnotemark[2] \\[2mm]
    $^1$01.AI, $^2$University of Waterloo,$^3$HKUST,\\ 
    $^4$University of Manchester, $^5$Multimodal Art Projection Research Community\\[2mm]
}

\begin{document}

\maketitle

\vspace{-0.7cm}
\begin{center}
    \url{https://github.com/Zheng0428/COIG-Kun}
\end{center}
\vspace{5pt}
\begin{abstract}
In this paper, we introduce Kun\footnote{The dataset is named Kun as Chinese pronunciation of Humpback~\cite{li2023self}.}, a novel approach for creating high-quality instruction-tuning datasets for large language models (LLMs) without relying on manual annotations. 
Adapting a self-training algorithm based on instruction back-translation and answer polishment, Kun leverages unlabelled data from diverse sources such as Wudao, Wanjuan, and SkyPile to generate a substantial dataset of over a million Chinese instructional data points. 
This approach presents a novel departure from traditional methods by using a self-curation process to refine and select the most effective instruction-output pairs. 
Our experiments with the 6B-parameter Yi model across various benchmarks demonstrate Kun's robustness and scalability.
Our method's core contributions lie in its algorithmic advancement, which enhances data retention and clarity, and its innovative data generation approach that substantially reduces the reliance on costly and time-consuming manual annotations. 
This methodology presents a scalable and efficient solution for improving the instruction-following capabilities of LLMs, with significant implications for their application across diverse fields.
\end{abstract}
\begin{figure*}[t]
  \centering
    \centering
    \includegraphics[width=\linewidth]{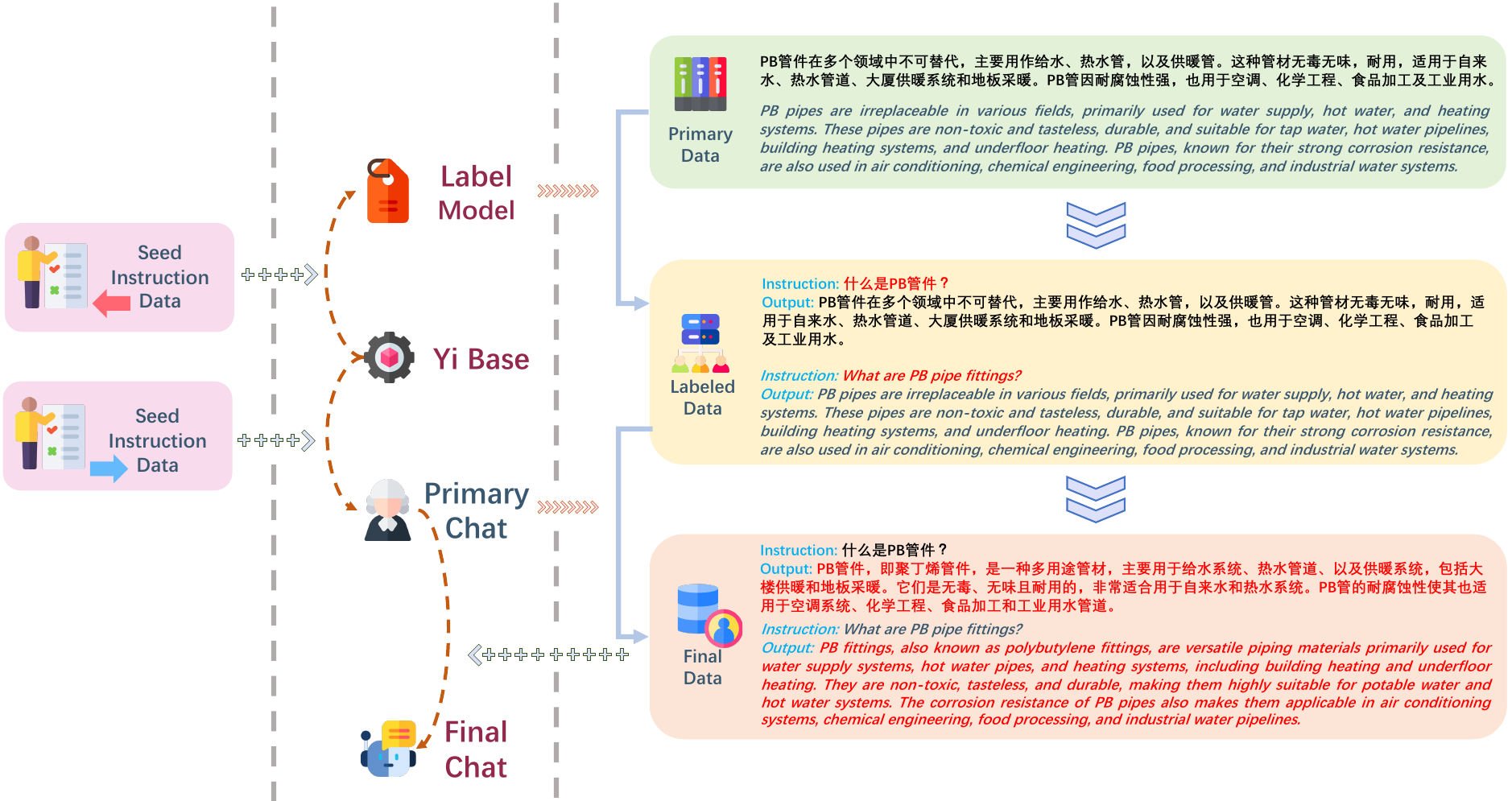}
  \hfill
    \centering
  \caption{Overview of \textit{Answer Polishment}.
Initially, the Yi base model is fine-tuned using quality seed instruction data to create a label and a primary chat model. The label model then annotates a large amount of primary data, turning it into labeled data. This is filtered and refined by rules and the primary chat model, producing the final dataset. This dataset is used to further train the primary chat model, resulting in an highly efficient final chat model.}
  \label{fig:flowchart}
\end{figure*}

\section{Introduction}

The development of large language models (LLMs) relies on human-annotated datasets, yet the creation of such datasets typically faces scalability issues due to the significant resources required. 
Our study introduces Kun, a novel approach leveraging unlabelled data to create a high-quality instruction-tuning dataset. 
This method diverges from manual annotations, employing a self-training algorithm that includes a unique process called AP (\textbf{A}nswer \textbf{P}olishment),

AP is central to Kun's strategy. 
It addresses a critical challenge in the Humpback~\citep{li2023self} method, where raw data, once labeled, are directly used in instruction datasets.
The unscreened raw data often mismatches between instructions and responses, as raw data may not inherently align with the instructional context. 
AP refines this raw data, ensuring a tighter correlation between the instructions and responses through a back-translation process. This leads to a dataset where each instruction-output pair is more coherent and contextually relevant.
Unlike methods dependent on LLMs~\citep{peng2023instruction,taori2023alpaca,zheng2023judging}, Kun offers an independent and scalable approach to instruction-based training.

We opt for the 6B-parameter Yi model due to its open-source nature and dependable performance\footnote{\url{https://github.com/01-ai/Yi}}. 
Its efficacy is tested and proven across diverse dataset sizes, including widely recognized benchmarks like C-EVAL~\citep{huang2023ceval} and CMMLU~\citep{li2023cmmlu}.
To evaluate the performance of the model, we design a comprehensive human evaluation which contains 500 prompts from ShareGPT-zh, covering various tasks. Responses generated by our model are compared with those from other models, showcasing the superiority of our \emph{Kun-52k} variant.
Further details can be found in \ref{sec:import area}.
Additionally, we evaluate the quality of our dataset, which includes 1,000 instruction-output pairs each from sources like Wudao~\citep{Wudao}, Wanjuan~\citep{he2023wanjuan}, and SkyPile~\citep{wei2023skywork}. This evaluation, focusing on clarity, feasibility, practicality, and alignment, ensures the high quality of our dataset.
The key contributions of our work are:

\begin{itemize}
\item \textit{Algorithmic Advancement}: AP in Kun enhances data retention and resolves ambiguities, leading to an expanded pool of high-quality data for fine-tuning.

\item \textit{Large-scale high quality data creation}: Over a million diverse Chinese instructional data points are produced from sources like Wudao, Wanjuan, and SkyPile, surpassing traditional crowdsourced annotations in quality and reducing reliance on manual annotation.
\end{itemize}

\section{Related Work}


\textbf{Instruction Tuning} 

Instruction tuning is widely recognized as a key technique for activating LLMs to adhere to human conversational norms.
~\citep{naturalinstructions,supernaturalinstructions,wang2023interactive}. 
Instruction tuning empowers various domain-specific or task-specific LLMs, including natural language generation evaluation~\citep{jiang2023tigerscore}, math~\citep{yue2023mammoth,xu2023wizardlm,azerbayev2023llemma}, code~\cite{luo2023wizardcoder}, music~\cite{li2023mertech,deng2023musilingo}, and medicine~\cite{wang2023huatuo}.
Instruction tuning not only tailors the models' task-specific responsiveness but also bolsters their cross-task generalization capabilities, thus enhancing performance across various dynamic application contexts~\citep{weifinetuned,sanh2022multitask,naturalinstructions,supernaturalinstructions}.
Recent studies have broadened the scope of instruction tuning to encompass a wider array of general tasks, notably incorporating input from users of language models~\citep{ouyang2022training,peng2023instruction}.

However, the open-source community is still lacking high-quality Chinese instruction tuning corpora. Current datasets, like COIG~\citep{zhang2023chinese}, BELLE~\citep{ji2023exploring}, MOSS~\citep{sun2023moss}, and OL-CC~\citep{olcc}, face issues such as limited scope, poor quality, commercial restrictions, or insufficient coverage. This gap hampers the advancement of LLMs in effectively processing and executing complex Chinese instructions, highlighting the critical need for more diverse and superior-quality datasets.


\textbf{Self-Improvement of LLMs}

\begin{wrapfigure}{r}{0.5\textwidth}
  \centering
  \includegraphics[width=0.5\textwidth]{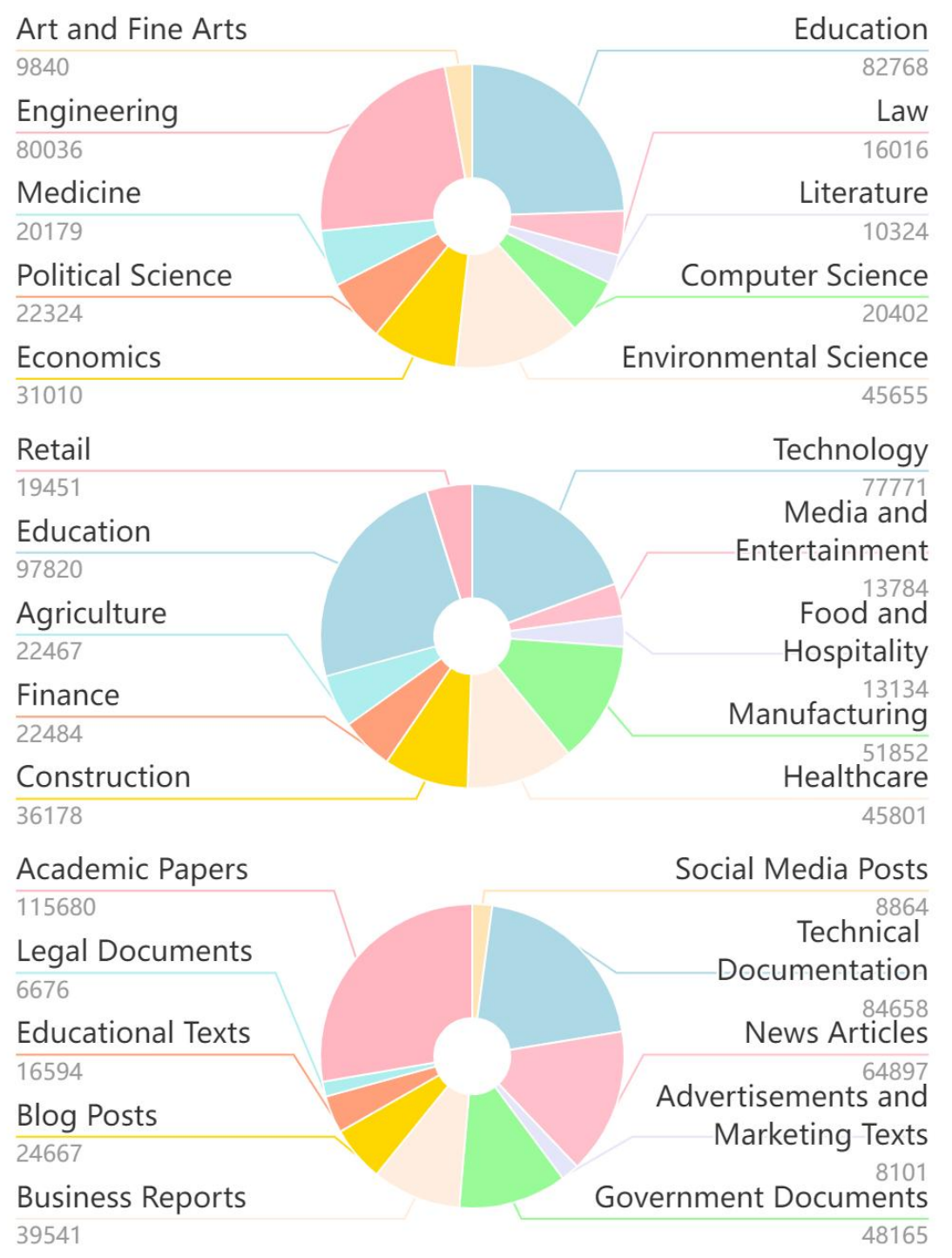}
  \caption{The top 10 categories in each of these three areas: Academic Disciplines, Industry Sectors, Text Type}
  \label{fig:type}
\vspace{-0.5cm}
\end{wrapfigure}

In fine-tuning LLMs, the availability of extensive, high-quality instructional data is crucial. Presently, the generation of such data mainly relies on human manual annotation, a labor-intensive method that lacks scalability for future data augmentation. 

An alternative approach involves deriving instructional data from more advanced LLMs~\citep{alpaca}, exemplified by extracting authentic instruction-response sets from dialogues within the GPT series models~\citep{selfinstruct}.
A more refined technique utilizes the Self-Instruction framework, autonomously generating additional instructional data from initial seed data. Combined with the Ada-Instruct or the Evol-Instruct framework, this approach can transform basic instructions into complex ones, specifically tailored for distinct tasks~\citep{cui2023ada,luo2023wizardcoder}.

Nevertheless, these instruction generation methodologies all require a robust teacher.The ultimate potential of the model is limited by the teacher's expertise or resource expenditure~\citep{li2023self}. 
To overcome this limitation, the SPIN~\citep{chen2024self} framework incorporates a self-play mechanism, It generates training data from previous iterations, refining its strategy by distinguishing between responses generated autonomously and those derived from human-annotated data. 

This gradual process elevates the LLM from a nascent model to a robust one. Considering the vast amount of knowledge present in web text, Humpback~\cite{li2023self} introduces a technique based on Instruction Backtranslation. 
This method allows a base model to independently utilize vast amounts of unlabeled data to generate a high-quality instruction tuning dataset. 
However, empirical findings indicate that the effectiveness of this method is still constrained by the seed model's performance and its ability to discern truly high-quality data.

\section{Method}

Our training methodology necessitates a foundational model, high-quality seed instruction data, and a substantial volume of unlabeled data, with the primary source being web text. Given the extensive content diversity inherent in large-scale web documents, which encompass a wide array of topics such as music, art, technology, etc., reflecting the broad spectrum of human interests, certain subsets of these documents may be more apt for generating instructions. Unlike labeled data, these documents lack predetermined outcomes or objectives.  This method involves the refinement and optimization of data selection during the fine-tuning process of Large Language Models (LLMs), as illustrated in Figure \ref{fig:flowchart}. This approach allows for the collection of a significant volume of instructional data at a low cost, circumventing the exorbitant expenses associated with manual labor, in a manner that is both academically rigorous and professional.
Our method consists of two main steps:

\begin{itemize}
\item  Supervised Fine-Tuning (\textbf{SFT}) with High-Quality Seed Data: This involves using SFT on the base model with high-quality seed data to create two models - the label model for annotating primary data and the primary chat for improving data quality.
\item  Quality Assessment and Refinement in Primary Chat: The primary chat assesses and refines the label model's output. This repeated process produces a lot of high-quality data, essential for the primary chat's further training. It leads to a high-performance final chat model, trained extensively with superior data.
\end{itemize}
A more comprehensive explanation of each step is provided subsequently.

\begin{figure}[ht]
  \centering
    \centering
    \includegraphics[width=0.48\textwidth]{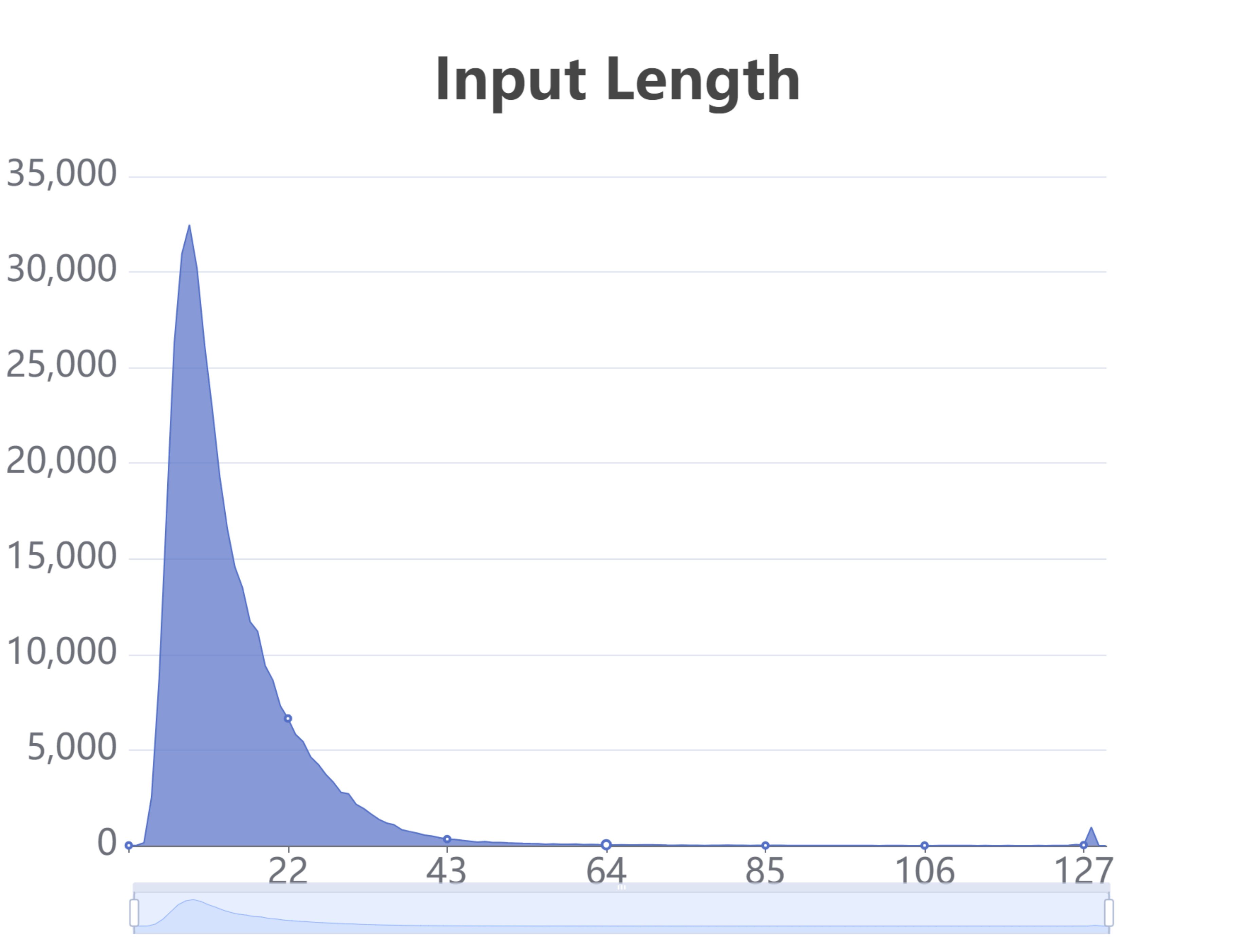}
  \hfill
    \centering
    \includegraphics[width=0.48\textwidth]{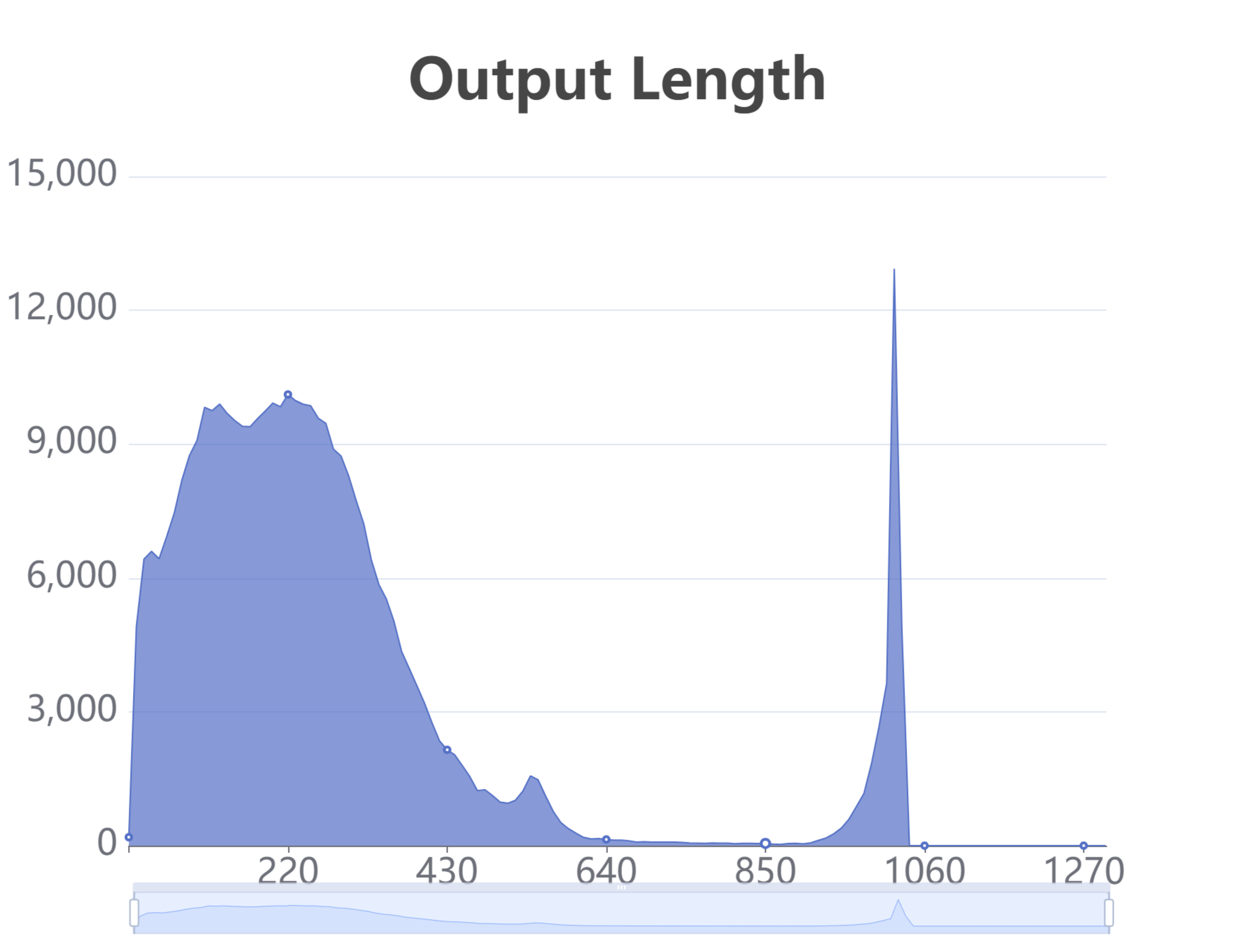}
  \caption{Length distribution of instructions and outputs based on Yi-6B model}
  \label{fig:length}
\end{figure}

\subsection{Initialization}

\textbf{Seed data.} We use 26,000 instructions and their corresponding outputs as our seed data. Each pair is hand-annotated and undergoes strict quality control to ensure high accuracy and consistency.

\noindent
\textbf{Primary Data.} The primary data originates from three significant Chinese datasets: WuDao, Wanjuan, and SkyPile. These datasets are distinguished by their extensive scale, diverse content, and comprehensive nature. Such characteristics make them ideal for mining golden texts aligned with specific user instructions. To facilitate this, we preprocess these datasets to isolate self-contained segments, denoted as ${y}_{i}$.

\noindent
\textbf{Supervised Fine-Tuning.} We utilize high-quality seed data to execute SFT on the foundational model, yielding two distinct models: the label model and the primary chat model.
\begin{itemize}

    \item Primary Chat Model: This model is fine-tuned utilizing the (instruction-output) pairs {(${x}_{i}$, ${y}_{i}$)} from the seed data. This process creates a forward model, $M_{xy} := p(y \mid x)$, with $x$ and $y$ maintaining their respective meanings as instructions and outputs.

   \item Label Model: Conversely, this model undergoes fine-tuning with the (output-instruction) pairs {(${y}_{i}$, ${x}_{i}$)} derived from the seed data, leading to the formation of a backward model, denoted as $M_{yx} := p(x \mid y)$. In this context, $x$ signifies the instruction, while $y$ denotes the corresponding output.
   
\end{itemize}

\subsection{Generating Candidate Instructions with the Label Model}

\begin{figure*}[ht]
  \centering
    \centering
    \includegraphics[width=\linewidth]{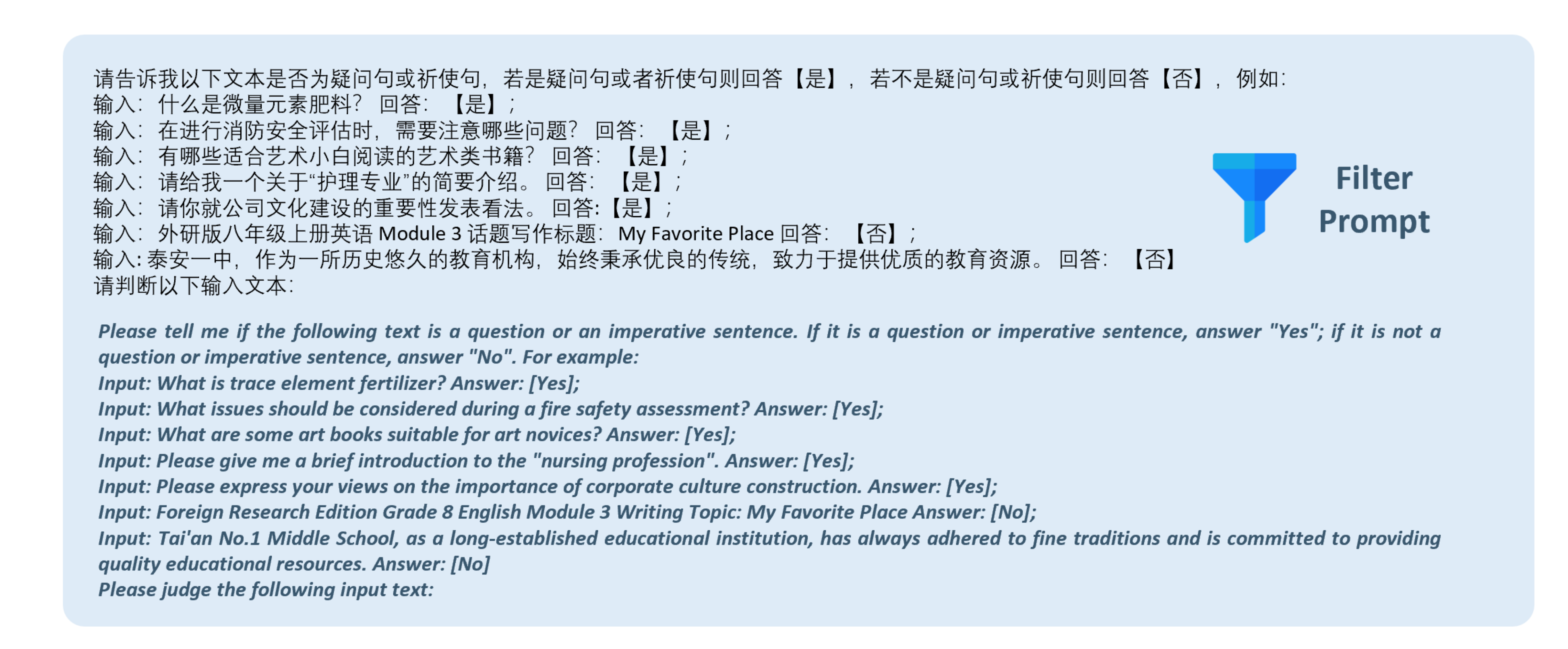}
  \hfill
    \centering
  \caption{Filter prompt we use to screen out unsuitable content for instructions.}
  \label{appendix:filter}
\end{figure*}

For each example ${y}_{i}$ in the unlabeled set, we utilize the backward model to infer a candidate instruction, denoted as $\hat{x}_{i}$. This procedure generates a collection of potential enhanced pairing data, represented as {($\hat{x}_{i}$, ${y}_{i}$)}. During annotation, initial filtering is based on perplexity (\textbf{ppl}) and length, and excludes any data exceeding 512 tokens. We also discard data unsuitable for instructional use, like purely descriptive statements, while retaining useful data, such as commands and questions. We apply a filter prompt in this selection, keeping only data that triggers a positive response. The final labeled dataset contains instruction and output components, with instructions from the label model and outputs from primary data, known as candidate labeled data. Figure \ref{appendix:filter} shows the filter prompt used in this stage.




\subsection{Refinement of Candidate Labeled Data Using the Primary Chat Model}

Given the extensive volume of primary data, not all candidate labeled data derived from it are of uniformly high quality. Our objective is to curate high-quality labeled data, necessitating rigorous filtration of the candidate labeled data.

Our experiments tested two different filtering methods, each with its strengths and weaknesses:
\begin{itemize}

    \item Comprehensive Scoring of Labeled Data: This method evaluates the full labeled data set based on a combined score, including instructions and outputs.

    \item Focused Scoring of Instruction Component: This technique only assesses the instruction part (output from the label model). High-scoring instructions are chosen, and then the output part of these selected data is refined.
\end{itemize}

Our analysis shows that the second method is more effective than the first. In the first method, good outputs are often discarded because of poor instructions from the label model, and the reverse is also true, causing unnecessary exclusions. Moreover, this approach occasionally  retains data with one poor quality instruction because the corresponding output is high quality, and vice versa, leading to uneven data quality and negatively impacting further training.

In contrast, the second method only scores the instruction component,
as in instruction tuning for LLM, instructions are often considered more important than outputs,Yet, it doesn't assess the output, sometimes leading to suitable instructions paired with unsuitable outputs. To address this, we use the primary chat model to evaluate and refine the instructions and outputs, ensuring they align well. This approach produces high-quality labeled data. The score and refine prompts we used in this process are shown in Figure \ref{appendix:score_and_refine}.



Utilizing the substantial volume of top-quality labeled data from these procedures, we further train the main chat model, achieving a high-performance final model, as shown in Experiments.


\section{Experiments}

In this section, we comprehensively detail the experimental procedures and methodologies employed in our study.

\subsection{Experimental Setup}

We first detail the experimental setup used in our study, covering the base model selection, fine-tuning process, baseline comparisons, and the evaluation methods.

\subsubsection{Base Model \& Finetuning}

In our experiments, we utilize the Yi model with 6B parameters\footnote{\url{https://huggingface.co/01-ai/Yi-6B}}, developed by 01.AI, as our foundational language model for fine-tuning. Renowned for its proficiency in both English and Chinese, the Yi series has shown impressive results on global benchmarks like the AlpacaEval Leaderboard~\citep{dubois2023alpacafarm,alpaca_eval} and SuperCLUE~\citep{xu2023superclue}.

The fine-tuning process is carried out using varying sizes of our high-quality, instructionally curated dataset. This phase is executed on a computing setup with 32 Nvidia A800 GPUs, amounting to a total of 192 GPU hours. We adopt a learning rate of 2e-5 and a batch size of 16, aiming for an optimal balance between computational efficiency and model performance. All the models have been fine-tuned with the same number of update steps.

\begin{table*}[htbp]
\begin{centering}
\small
\begin{tabular*}{\textwidth}{@{\extracolsep{\fill}}lccccccc}
\toprule[1pt] 
\multirow{2}{*}{ \textbf{Source} } 
& \multicolumn{3}{c}{ \textbf{Instruction Quality} } & \multicolumn{3}{c}{ \textbf{Output Quality} }  \\
 & Clarity\% & Feasibility\% & Practicality\% & Excellent\% & Pass\% & Fail\%  \\
\midrule[1pt] 
\textbf{Wudao}        & 96.67          & 96.40           & 96.87            & 69.50           & 20.03              & 10.47    
\\
\textbf{Wanjuan}    & 98.27          & 97.63           & 96.57            & 85.63           & 11.13              & 3.24    
\\
\textbf{Skypile}      & 98.90          & 98.37           & 95.40            & 42.73           & 40.43              & 16.84    
\\ \midrule 
\textbf{ALL}     & 97.94          & 97.47            & 96.28            & 66.00           & 23.87              & 10.13    
\\
\bottomrule[1pt]
\end{tabular*}
\end{centering}
\caption{Manual Quality Analysis of Synthetic Data Generated by Kun.}\label{tab:statistics}
\end{table*}
\subsubsection{Baselines}

For the Kun dataset, we annotated command data from three sources: Wudao, Wanjuan, and Skypile. Quantitative details of this augmented dataset are provided in Figure \ref{fig:quantity}. In evaluating the performance of Kun, our study contrasts it with data curated from four prominent Chinese open-sourced datasets, including COIG~\citep{zhang2023chinese,coig-pc,coig-pc-lite}, OL-CC~\citep{olcc}, and BELLE~\citep{ji2023exploring}. These datasets are unique in their composition and focus, providing a comprehensive basis for comparison. 





\subsubsection{Evaluation}
\label{sec: model evaluation}
\noindent
\textbf{Human Evaluation.} To assess the general quality of model responses, we conduct human evaluations using a test set of 500 prompts sampled from ShareGPT-zh. These prompts, derived from real world user inquiries, encompass a diverse array of tasks, such as creative writing, information seeking, providing guidance, logical reasoning, storytelling, problem-solving, etc.

For the evaluation, responses generated by different models for each prompt are presented side-by-side. Human evaluators are asked to choose their preferred answer, providing a direct comparison of model performance. In total, eight models were compared.For this evaluation, we engage a team of experienced crowdsource annotators, ensuring a balanced and unbiased assessment. Detailed examples that show the comparison process can be found in Figure \ref{appendix:example}.
\noindent
\textbf{Standard Benchmarks.} In addition to human evaluations, the models are also assessed using two standard benchmarks for Chinese LLM evaluation: C-EVAL ~\citep{huang2023ceval} and CMMLU ~\citep{li2023cmmlu}.

These evaluation methods, comprising both human judgment and standardized benchmarks, offer a multifaceted perspective on the capabilities of the Kun model, enabling a thorough comparison with existing models and datasets.

\subsection{Augmentation Data Statistics}

In this section, we delve into the detailed statistical analysis and diversity assessment of our augmented dataset, as well as the rigorous quality evaluation conducted using human annotators. 
By exploring the comprehensive scale, varied nature, and assessed quality of the instruction-output pairs, we aim to highlight the robustness and reliability of the data curated for our study.

\subsubsection{Statistics and Diversity}

Our work involve the purification of approximately 377,592 high-quality instruction-output pairs from the Wudao, Wanjuan, and Skypile datasets.We analyze a 10\% subset of instructions from the past 20 years, revealing significant temporal diversity with 56\% of instructions from the recent three years (Figure \ref{fig:year}). The variation in instruction and output lengths, analyzed using the Yi-6B model, is shown in Figure \ref{fig:length}, reflecting content complexity.

To assess instruction diversity, we categorize them into 24 academic disciplines, 16 industry sectors, and 15 text types as per Wikipedia\footnote{\url{https://www.wikipedia.org}} using the Qwen-72B-Chat\footnote{\url{https://huggingface.co/Qwen/Qwen-72B-Chat}}~\citep{bai2023qwen}. Repeated for accuracy, this categorization highlights the data's range, as shown in Figure \ref{fig:type}, where the top 10 categories in each area signify its broad scope.

\begin{figure}[htbp]
  \centering

  \begin{minipage}{0.4\textwidth}
    \centering
    \begin{tabular}{lc}
      \toprule[1pt] 
      \textbf{Source} & \textbf{Consistency\%} \\
      \midrule[1pt] 
      Clarity         & 96.87 \\
      Feasibility     & 97.73 \\
      Practicality    & 97.43 \\
      \midrule 
      ALL             & 92.13 \\
      \bottomrule[1pt]
    \end{tabular}
    \captionof{table}{The proportion of identical evaluations from three assessors on a single dimension. \textbf{All}: The proportion of consistent assessments across all three dimensions within the same item.}
    \label{tab:Consistency}
  \end{minipage}
  \hfill
  \begin{minipage}{0.50\textwidth}
    \centering
    \includegraphics[width=0.9\linewidth]{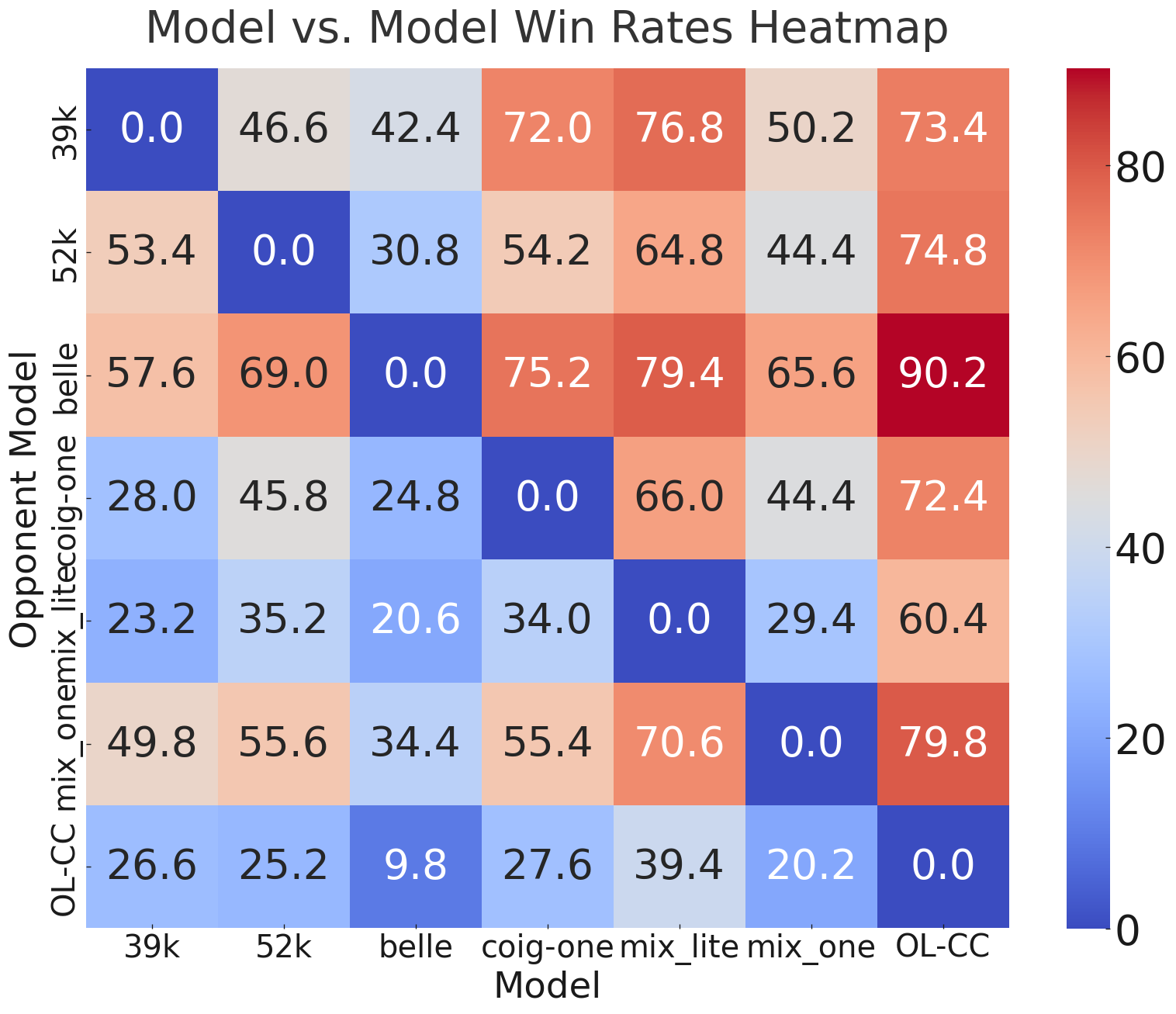}
    \caption{Heatmap of Model Comparative Win Rates in Human Evaluations. mix-lite:Kun-26k+COIG-lite.mix-one:Kun-26k+COIG-one}
    \label{fig:heatmap}
  \end{minipage}
  \vspace{-0.5cm}
\end{figure}

\subsubsection{Quality Evaluation}
\label{sec: data evaluation}

A critical aspect of our dataset curation process is the rigorous data quality assessment. We conduct a comprehensive quality evaluation of the instruction-output pairs to achieve this. For augmented data curated from each source (Wudao, Wanjuan, and Skypile), we randomly select 1,000 instruction-output pairs, resulting in 3,000 pairs subjected to an independent quality assessment. 

\noindent
\textbf{Instruction Quality.} For instruction quality, a team of 24 professional annotators with a bachelor's degree or higher evaluates each instruction across three key dimensions: clarity, feasibility, and practicality. Each aspect is assessed with a simple yes/no answer, providing a straightforward yet effective measure of instruction quality. The evaluation criteria are as follows:
\begin{itemize}
    \item \textbf{Clarity:} Evaluators determine whether the instruction was unambiguous and coherent, encompassing necessary information without any vague terms or explanations.

    \item \textbf{Feasibility:} Evaluators assess whether the instruction was valid and answerable within the context and scope of the model's capabilities.

    \item \textbf{Practicality:} Evaluators judge the relevance of the instruction in everyday scenarios.
\end{itemize}

\noindent
\textbf{Output Quality.} The quality of the outputs is evaluated based on their alignment with the instructions. Evaluators are asked to rate each output as \emph{Excellent}, \emph{Pass}, or \emph{Fail}, based on how well it met the requirements and intent of the instruction.

\begin{figure}[ht]
    \centering
    \begin{minipage}{0.45\textwidth}
        \centering
        \vspace{0.2cm}
        \includegraphics[width=\textwidth]{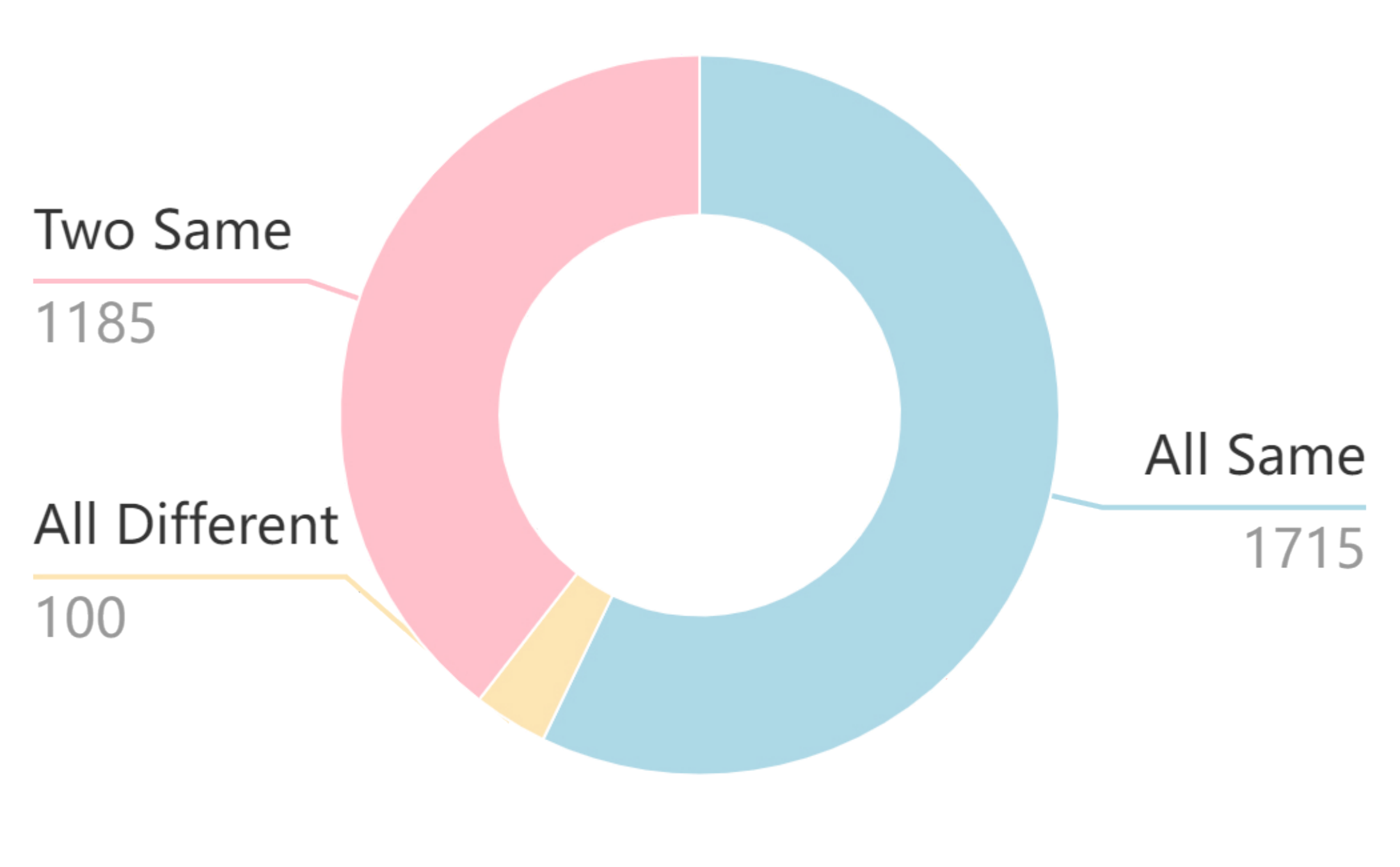}
        \caption{Distribution of evaluation identical ratings for instruction-response from three evaluators.}
        \label{fig:Consistency}
    \end{minipage}
    \hfill
    \begin{minipage}{0.45\textwidth}
        \centering
        \vspace{-0.1cm}
        \includegraphics[width=\textwidth]{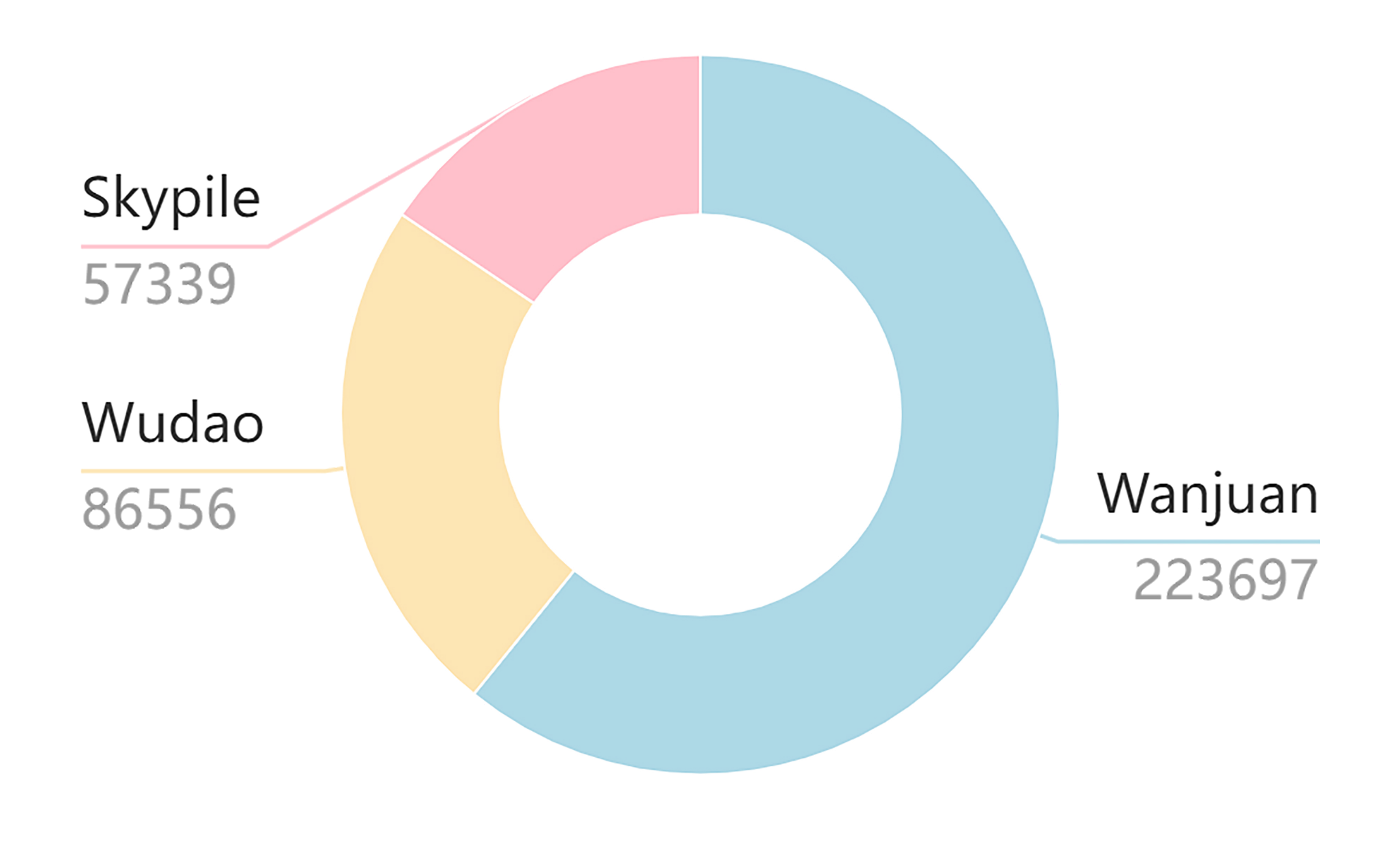}
        \vspace{0.03cm}
        \caption{Distribution of Data Sources}
        \label{fig:quantity}
    \end{minipage}
\end{figure}

To ensure objectivity and reliability, three different evaluators evaluate each instruction-output pair.  The consistency rates for the evaluation across the three dimensions of the instructions have all exceeded 90\%, and the evaluation of the instruction-response are also Consistently. This results demonstrate a significant degree of consistency in their judgments. Further details on evaluating identical ratings are presented in Figure \ref{fig:Consistency} and Table \ref{tab:Consistency}. Examples that demonstrate the process of assessing can be found in Figure \ref{appendix:example}
﻿


As indicated in Table \ref{tab:statistics}, the instruction quality across all sources is consistently high, suggesting effective formulation and clarity in conveying their purposes. However, the output quality varies more noticeably among the sources. While some sources like Wanjuan exhibite a high percentage of "Excellent" outputs, others such as Skypile demonstrate a more diverse distribution of output quality. This section presents the analysis of our experiment results, encompassing human evaluation outcomes and performance on standard benchmarks.

\subsubsection{Human Evaluation}
\label{sec:import area}

Our human evaluation results are illustrated through the heatmap in Figure \ref{fig:heatmap}, which shows the model vs. model win rates, with color variations indicating the relative performance strengths. 
The heatmap highlights that the \emph{Kun-52k} model emerges as the most dominant, followed by the mixed model, showcasing their superior ability to handle a wide range of prompts. 
In contrast, the baseline models \emph{COIG-39k}, and \emph{Belle-52k}, garner lower preference percentages. This suggests that despite their strengths, these models may not align as closely with user expectations or prompt requirements as the Kun models.
Further analysis is provided in Appendix A.4.

\subsubsection{Standard Benchmarks}
The performance of the models on standard benchmarks, Table \ref{tab:benchmark_stats} presents the performance statistics of various models on the C-EVAL and CMMLU benchmarks. As shown, we evaluate numerous models, including different sizes of the Kun model, baseline models, and mixed models. Each model's performance is measured in terms of perplexity and generation quality, providing a comprehensive view of its strengths and weaknesses.

From the table, we observe that \emph{Kun-39k} generally exhibits lower perplexity and higher generation quality, confirming its top-tier performance in language understanding and generation. Interestingly, the mixed model display robust performance, with the mixed model often outperforming \emph{Kun-52k}. The baseline models and smaller Kun variants present mixed results, excelling in some metrics while falling short in others. 
These highlight potential areas for further improvement in model training and fine-tuning strategies.

\begin{table}{}
\centering
\small
\begin{tabular}{lcccc}

\toprule 
\multirow{2}{*}{\textbf{Model}} & \multicolumn{2}{c}{\textbf{C-EVAL}} & \multicolumn{2}{c}{\textbf{CMMLU}} \\
                                & PPL    & GEN   & PPL    & GEN \\

\midrule

Kun-9k                          & 72.95               & 72.96             & 73.53               & 24.53    \\
Kun-26k         &73.73     &\underline{73.26}           & \definecolor{lightblue}{rgb}{0.68, 0.85, 0.9}\colorbox{lightblue}{\underline{75.75}}               & 25.23    \\
Kun-39k                         & \definecolor{lightblue}{rgb}{0.68, 0.85, 0.9}\colorbox{lightblue}{\underline{73.84}}               & \definecolor{lightblue}{rgb}{0.68, 0.85, 0.9}\colorbox{lightblue}{\textbf{73.38}}             & 75.68               & 25.31    \\
Kun-52k                         & 73.36               & 72.72             & 75.48               & 25.50    \\
Kun-100k                        & 72.95               & 72.96             & 73.53               & 24.53    \\
Kun-200k                        & 73.36               & 72.07             & 74.81               & 31.81    \\
Kun-360k                        & 73.25               & 71.88             & 74.14               & \definecolor{lightblue}{rgb}{0.68, 0.85, 0.9}\colorbox{lightblue}{\textbf{46.58}}    \\
\midrule
Kun-52k + & & & &\\
COIG-one-38k          & \multirow{-2}{*}{73.61}               &\multirow{-2}{*}{\textbf{73.38}}     & \multirow{-2}{*}{75.70}     & \multirow{-2}{*}{34.84}    \\
\midrule
COIG-38k                & \textbf{74.09}               & 72.95             &\textbf{75.92}               & 38.40    \\
Belle-52k                       & 72.31               & 72.56             & 74.73               &\underline{44.56}    \\
OL-CC-10k                       & 72.00               & 71.62             & 74.97               & 34.38    \\
\bottomrule
\end{tabular}

\caption{Performance Statistics on Standard Benchmarks.The best results in
each section are \textbf{Bold} , the second-best results are \underline{underlined}, while the results of our best model are in \definecolor{lightblue}{rgb}{0.68, 0.85, 0.9}\colorbox{lightblue}{Blue}. PPL:Perplexity  GEN:Generation }

\label{tab:benchmark_stats}
\end{table}

\section{Conclusion}
Our approach represents a breakthrough in instruction-tuning for LLMs, utilizing a novel self-training algorithm to leverage over a million quality Chinese instructional data points from diverse sources effectively. 
This strategy, different from manual annotations, 
not only enhances the instruction-following capabilities of LLMs but also ensures the high quality and diversity of training data. 
Empirical evaluations using the 6B-parameter Yi model across benchmarks like C-EVAL, CMMLU, and human evaluations, have demonstrated its robustness and scalability. 
Innovations within our approach, such as AP, have notably improved data retention and clarity, offering a scalable, efficient way to augment LLMs' instructional capabilities. This research not only progresses the field but also broadens LLMs' application scope, offering a novel solution to the challenges in developing instruction-tuning datasets.

\section{Ethics Statement}
This study adheres to the highest ethical standards, ensuring all research activities are conducted with a commitment to responsibility and respect for participant rights. Our ethical policy encompasses data usage, intellectual property rights respect, and research transparency. To safeguard data privacy and security, particularly when handling unlabeled data from sources like Wudao, Wanjuan, and SkyPile, we implement stringent measures to comply with data protection laws, especially concerning personal information. This involves anonymizing and desensitizing data prior to utilization. In terms of intellectual property rights, we ensure that all employed data and generated guiding points adhere to applicable copyright laws and intellectual property agreements, thereby avoiding infringement on any third-party intellectual property rights. Moreover, we pledge to provide a comprehensive account of our research methodology, detailing the processes of data generation, model training, and evaluation, to facilitate reproducibility and validation by the academic community.

\section{Limitations}
This study introduces an innovative methodology for generating data to reduce the reliance on costly and time-consuming manual annotations. However, there are limitations impacting the generalizability and scalability of our findings. 
Firstly, the diversity of data: the methodology, while capable of generating instructional data from a vast pool of unlabeled data, may produce data whose quality and diversity are constrained by the original data source's breadth and caliber. This is particularly relevant when generating instructions for niche or specialized domains, where additional methods might be necessary to guarantee comprehensive coverage and precision. Secondly, the generalization capability of the model: although tests on the 6B-parameter Yi model confirm the methodology's efficacy, its performance and applicability could differ among models of various sizes and tasks. Its effectiveness might require further investigation, especially in smaller models or those not designed for Chinese language processing. Thirdly, the assumptions underlying the algorithm: the study's self-training algorithm relies on instruction back-translation and answer embellishment. These premises might not hold across all instruction or answer types, notably for tasks demanding specialized knowledge or creativity, where the generated instructional data may not adequately reflect complex cognitive processes. Future Directions: Subsequent research should examine the methodology's applicability to large-scale language models across different languages and the generation of high-quality instructional data in specialized fields such as medicine or law. Additionally, advancing self-training algorithms to more effectively manage intricate and specialized instructions represents a crucial avenue for exploration.

\bibliography{colm2024_conference}
\bibliographystyle{colm2024_conference}

\clearpage
\appendix
\section{Appendix}

\subsection{Examples in Human Evaluation of Data and Model Performance} 

\begin{figure*}[ht]
  \centering
    \centering
    \includegraphics[width=0.91\linewidth]{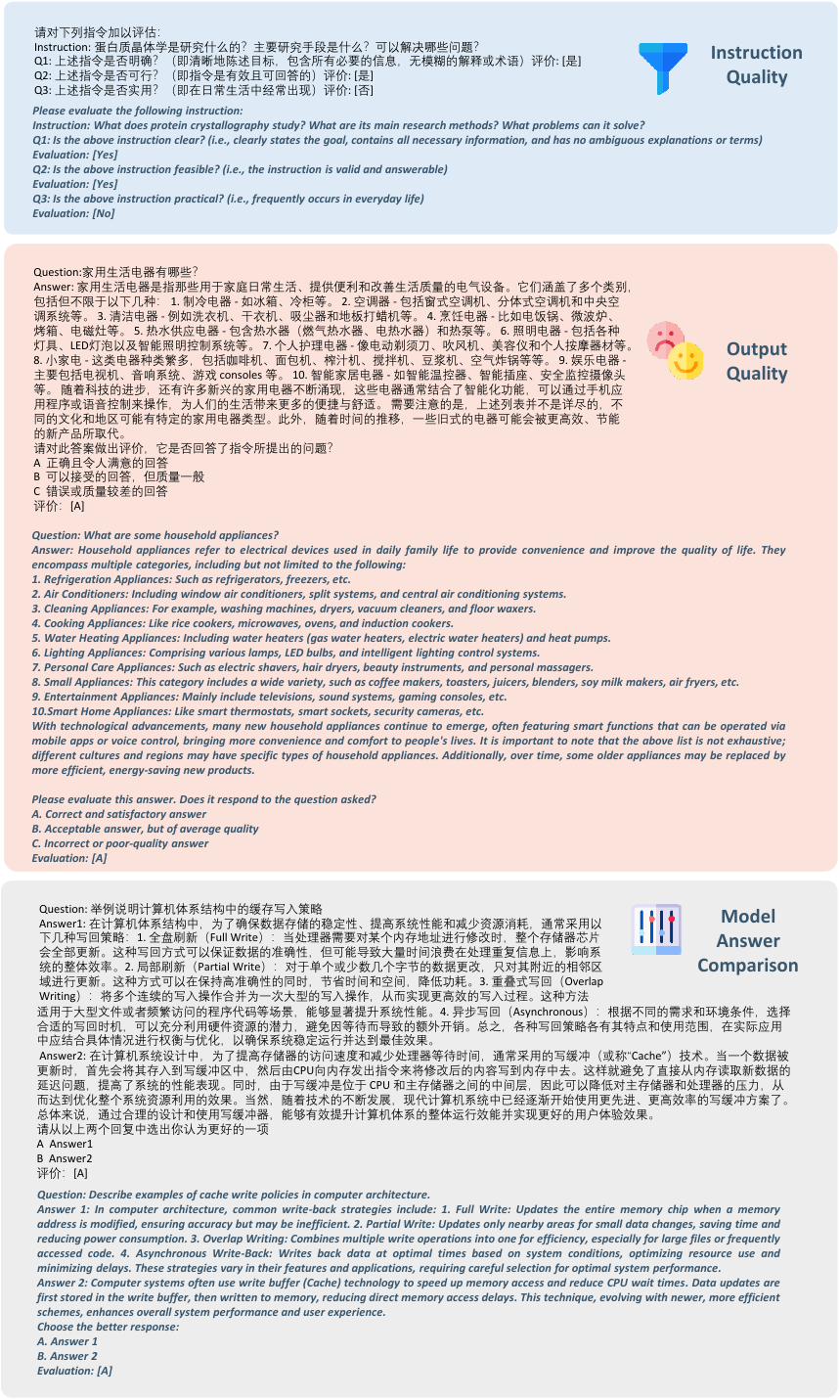}
  \hfill
    \centering
  \caption{Examples in Human Evaluation of Data and Model Performance.}
  \label{appendix:example}
\end{figure*}

Figure \ref{appendix:example} illustrates specific examples from our human evaluation process, as detailed in Sections \ref{sec: data evaluation} and \ref{sec: model evaluation}. This figure includes the evaluation of both instruction and output quality, along with a comparison of model answers. For each example, the original question is provided, followed by the evaluated responses. 

The instruction quality is assessed for clarity, feasibility, and practicality to evaluate the precision in formulating instructions. The output quality assessment focuses on the extent to which each response meets the instruction's requirements. Additionally, the model comparison examines which answers align most closely with human preferences, highlighting the practical effectiveness of the models under consideration.

\subsection{Data Filtering Rules} 
The text provided lists the manually established rules we use to delete some low-quality data in the process of generating data to improve data quality. These rules are used to filter the generated instructions or responses in cases with very obvious low-quality characteristics, thereby ensuring the reliability of the entire process.
1.Sensitive information such as phone numbers, home addresses, etc.
Further analysis is provided in Figures \ref{fig:mix_52k}, which depict the human preference evaluation win rates of \emph{Kun-52k} and the mixed model, respectively, when compared with other models. Notably, the mixed model demonstrates a higher win rate than \emph{Kun-52k}, indicating its enhanced effectiveness in meeting human evaluators' preferences.
\begin{itemize}
   \item Paragraphs that are largely repetitive.
   \item The length is too short (length <= 4).
   \item The text contains a large number of meaningless characters.
   \item Contains specific low-quality keywords.
   \item Refusal to answer.
   \item Text formatting errors.
\end{itemize}

\subsection{Years}
\begin{figure}[h]
  \centering
    \centering
    \includegraphics[width=0.5\linewidth]{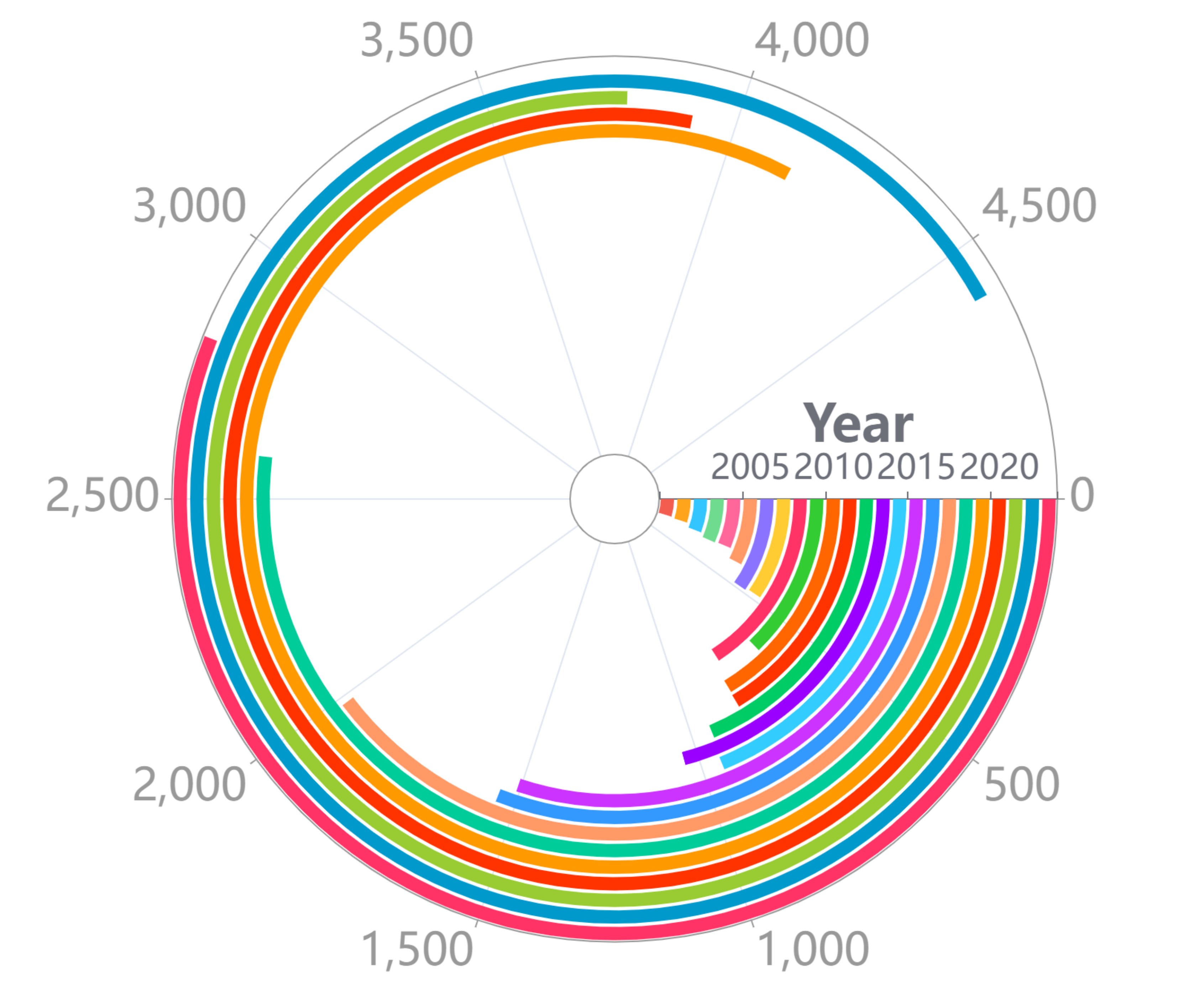}
  \hfill
    \centering
  \caption{Instructions spanning over 20 years, with 56\% from the last five years.}
  \label{fig:year}
\end{figure}
\subsection{More showcases of human evaluation results}

\begin{figure*}[htbp]
    \centering
    \begin{minipage}{0.40\textwidth}
        \centering
        \includegraphics[width=\textwidth]{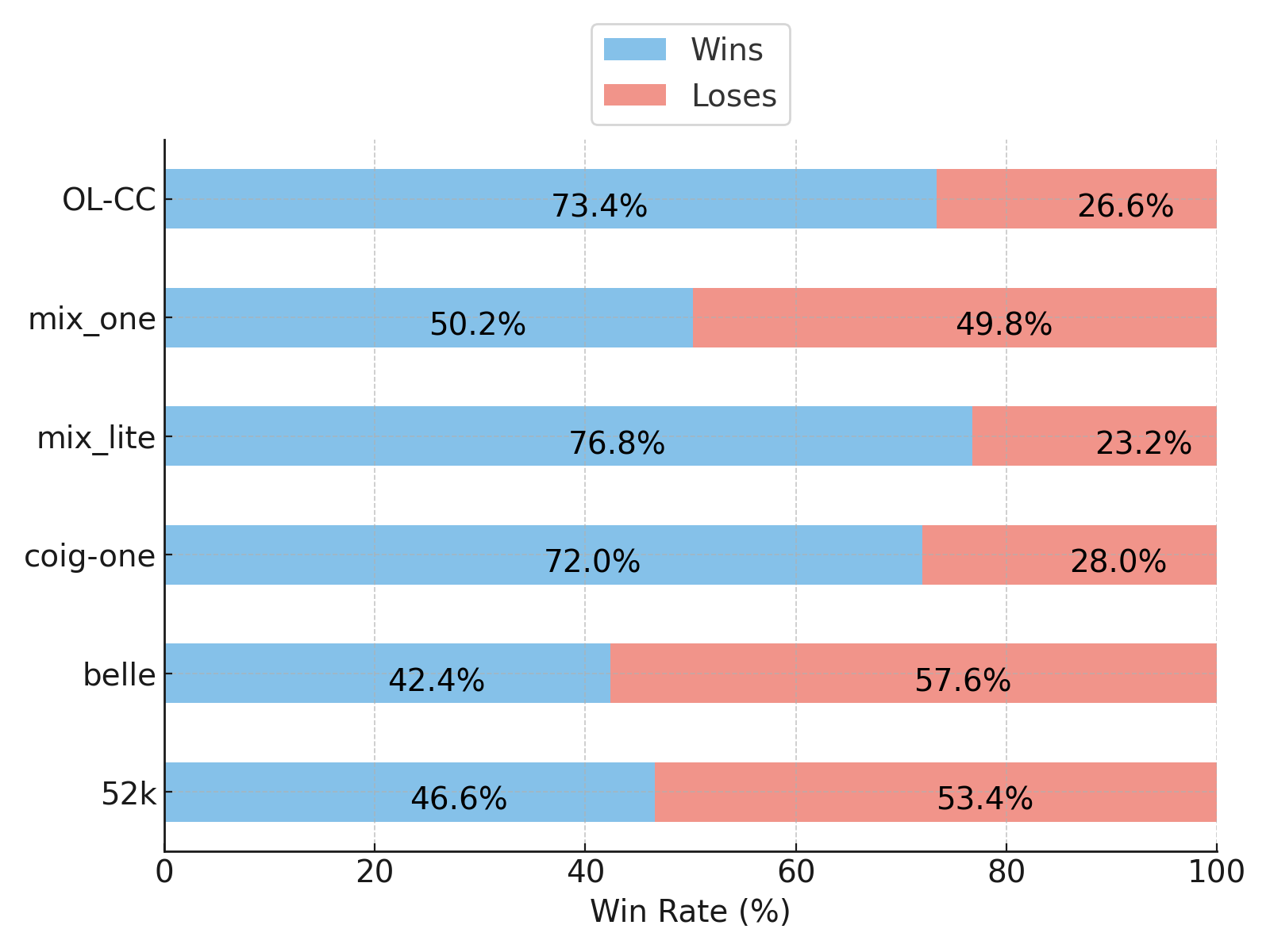} %

        \label{fig:52k}
    \end{minipage}
    \hfill
    \begin{minipage}{0.40\textwidth}
        \centering
        \includegraphics[width=\textwidth]{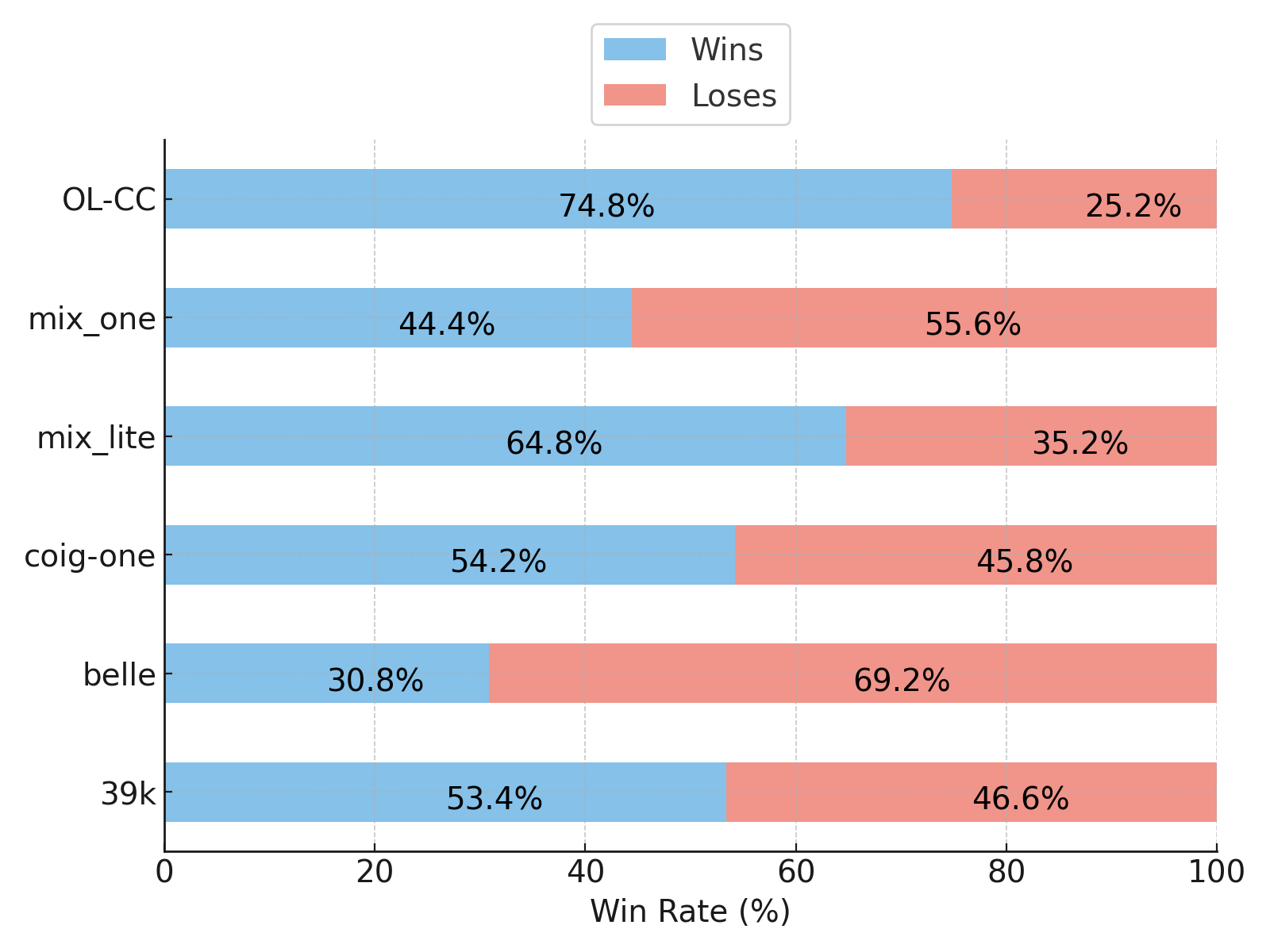}
        \label{fig:mix}
    \end{minipage}
    \hfill
    \begin{minipage}{0.40\textwidth}
        \centering
        \includegraphics[width=\textwidth]{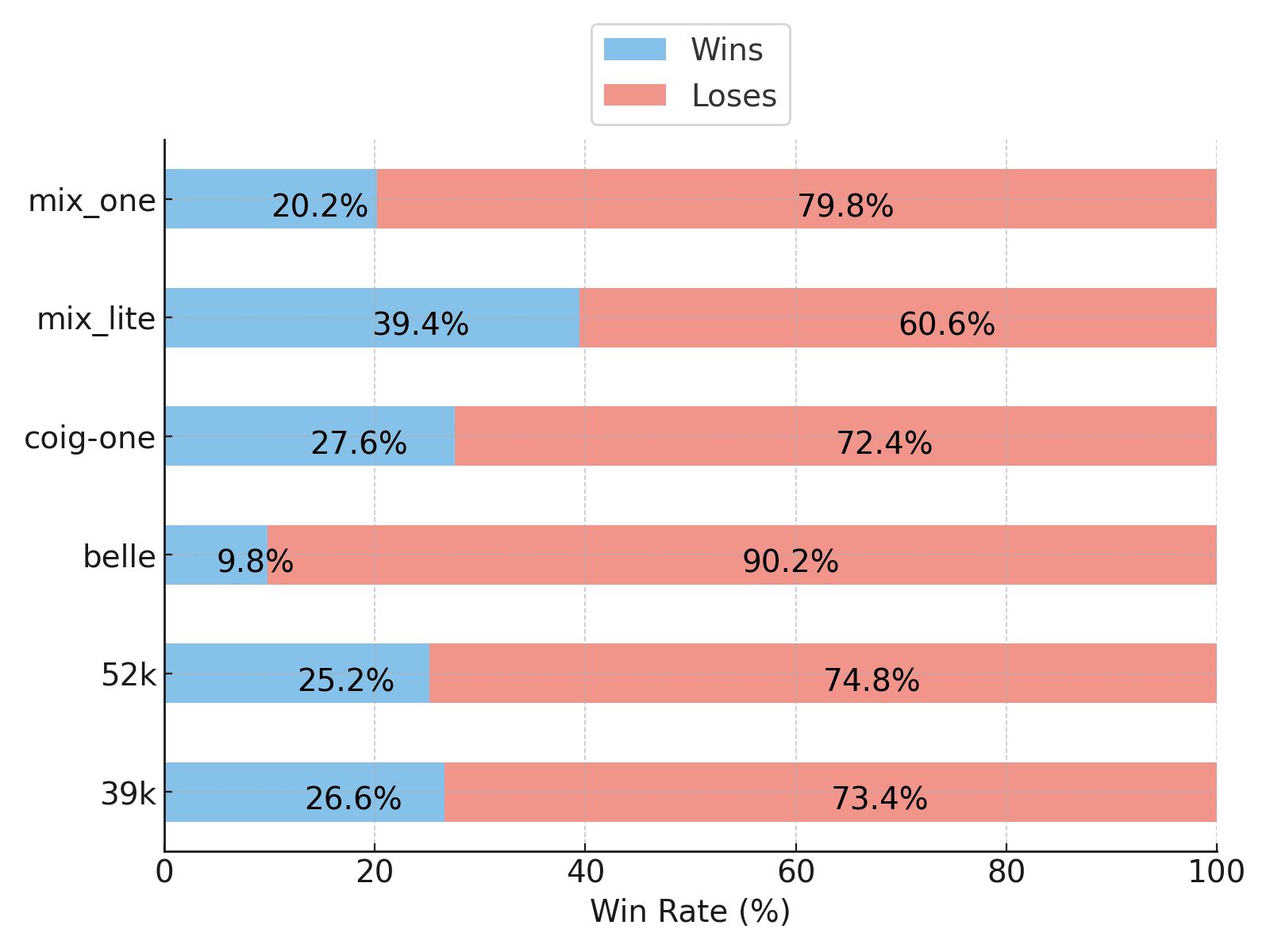}
        \label{fig:mix}
    \end{minipage}
    \hfill
    \begin{minipage}{0.40\textwidth}
        \centering
        \includegraphics[width=\textwidth]{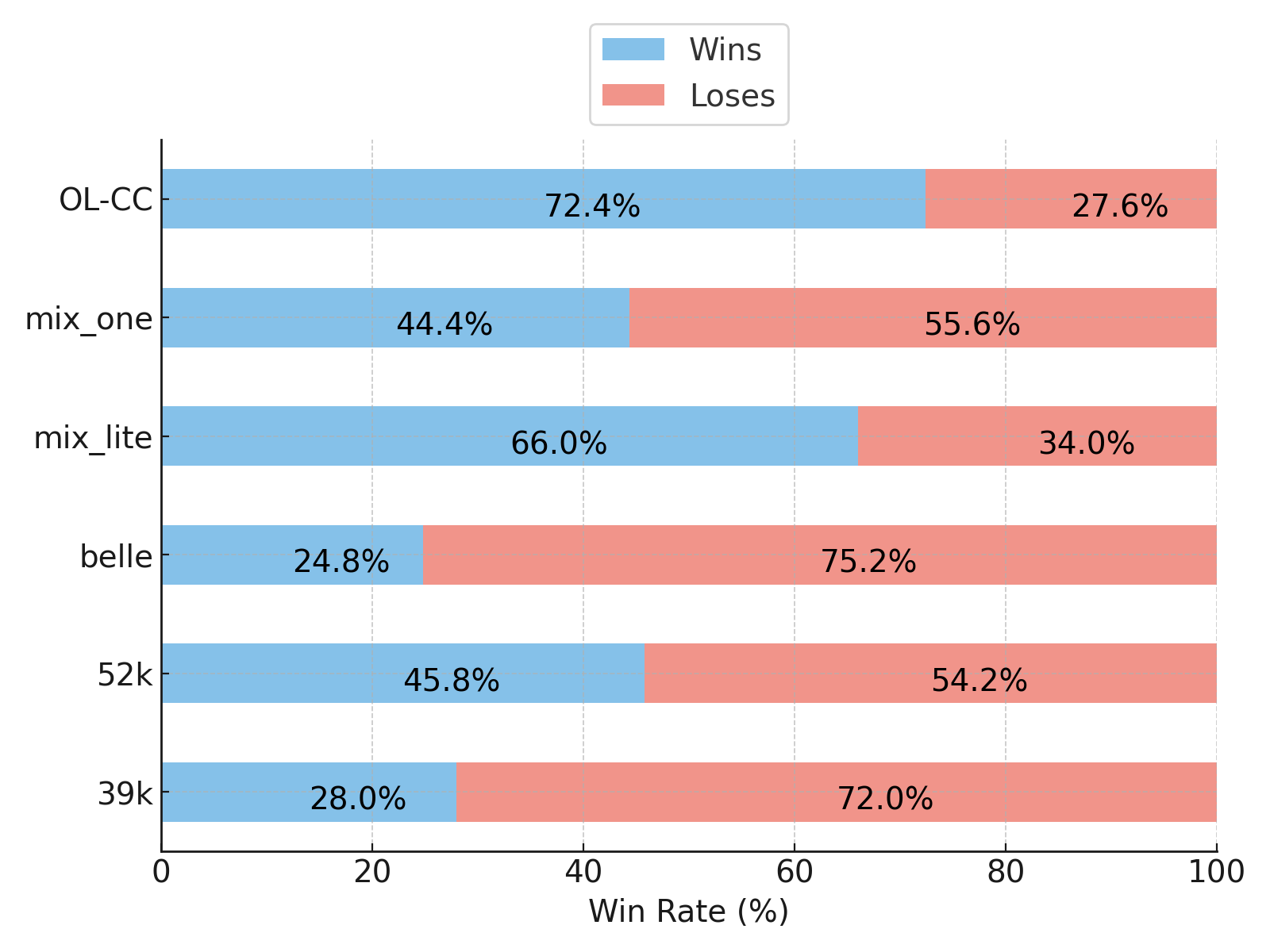}
        \label{fig:mix}
    \end{minipage}
    \hfill
    \begin{minipage}{0.40\textwidth}
        \centering
        \includegraphics[width=\textwidth]{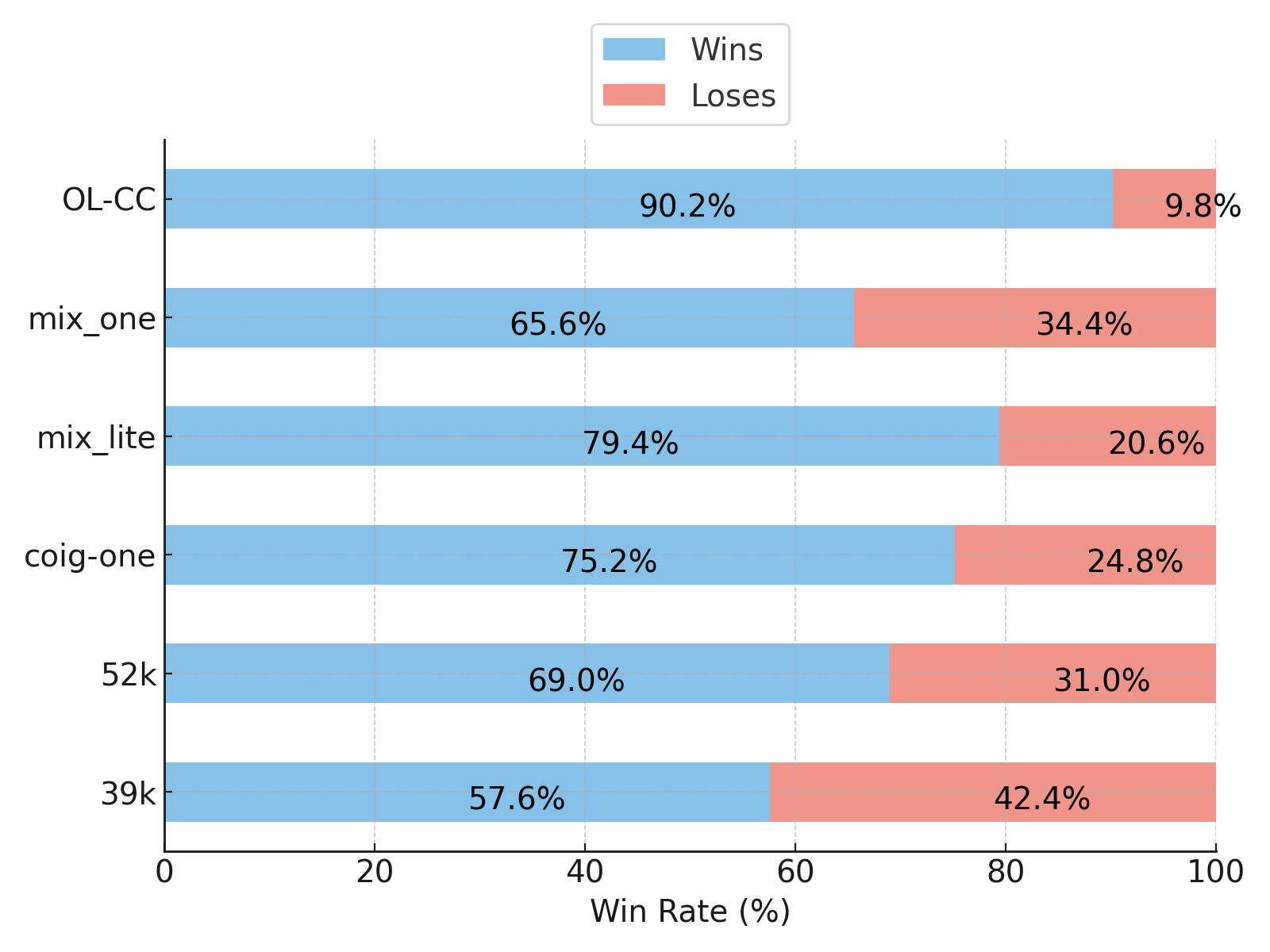}
        \label{fig:mix}
    \end{minipage}
    \hfill
    \begin{minipage}{0.40\textwidth}
        \centering
        \includegraphics[width=\textwidth]{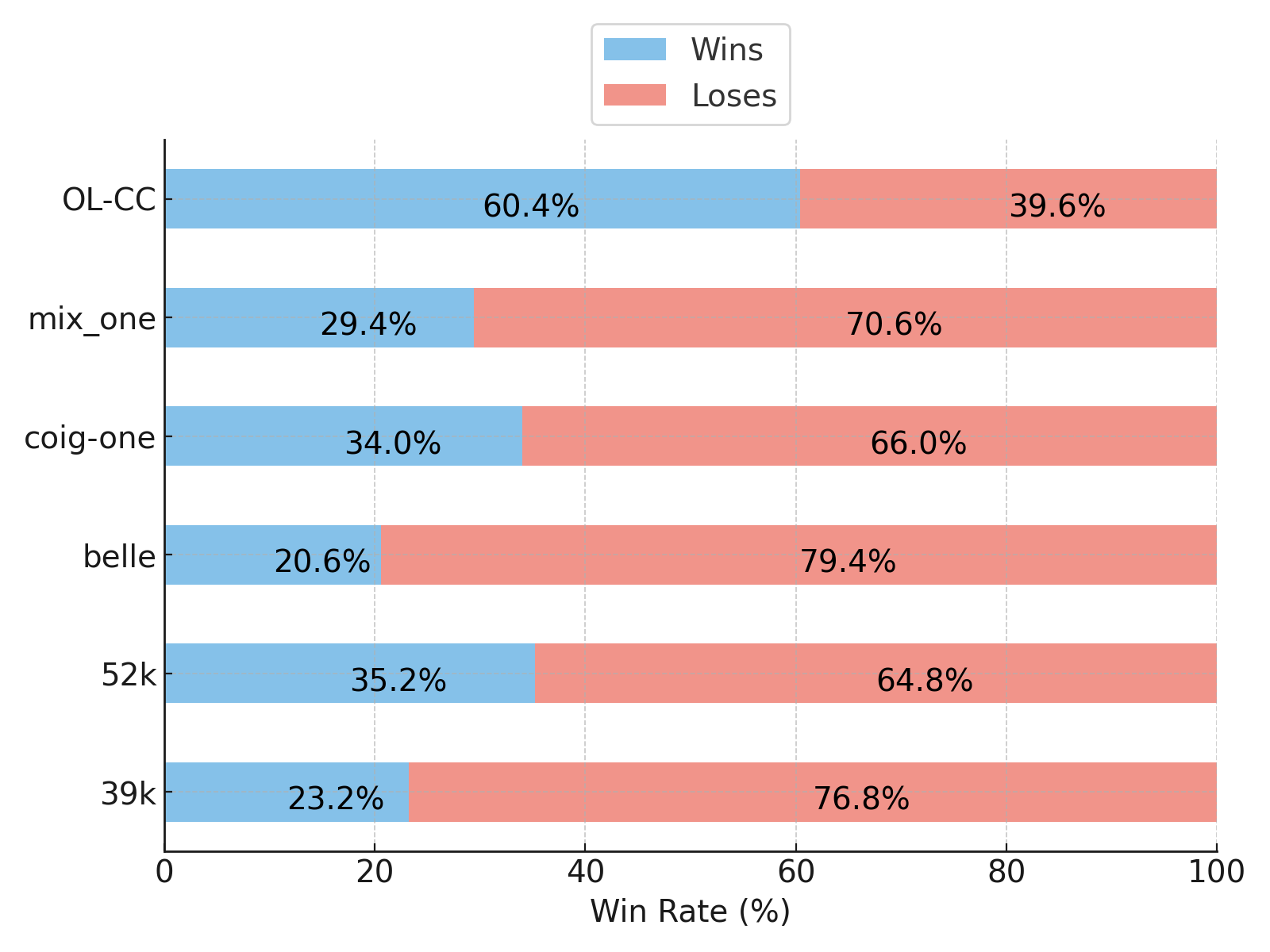}
        \label{fig:mix}
    \end{minipage}
    \hfill
    \begin{minipage}{0.40\textwidth}
        \centering
        \includegraphics[width=\textwidth]{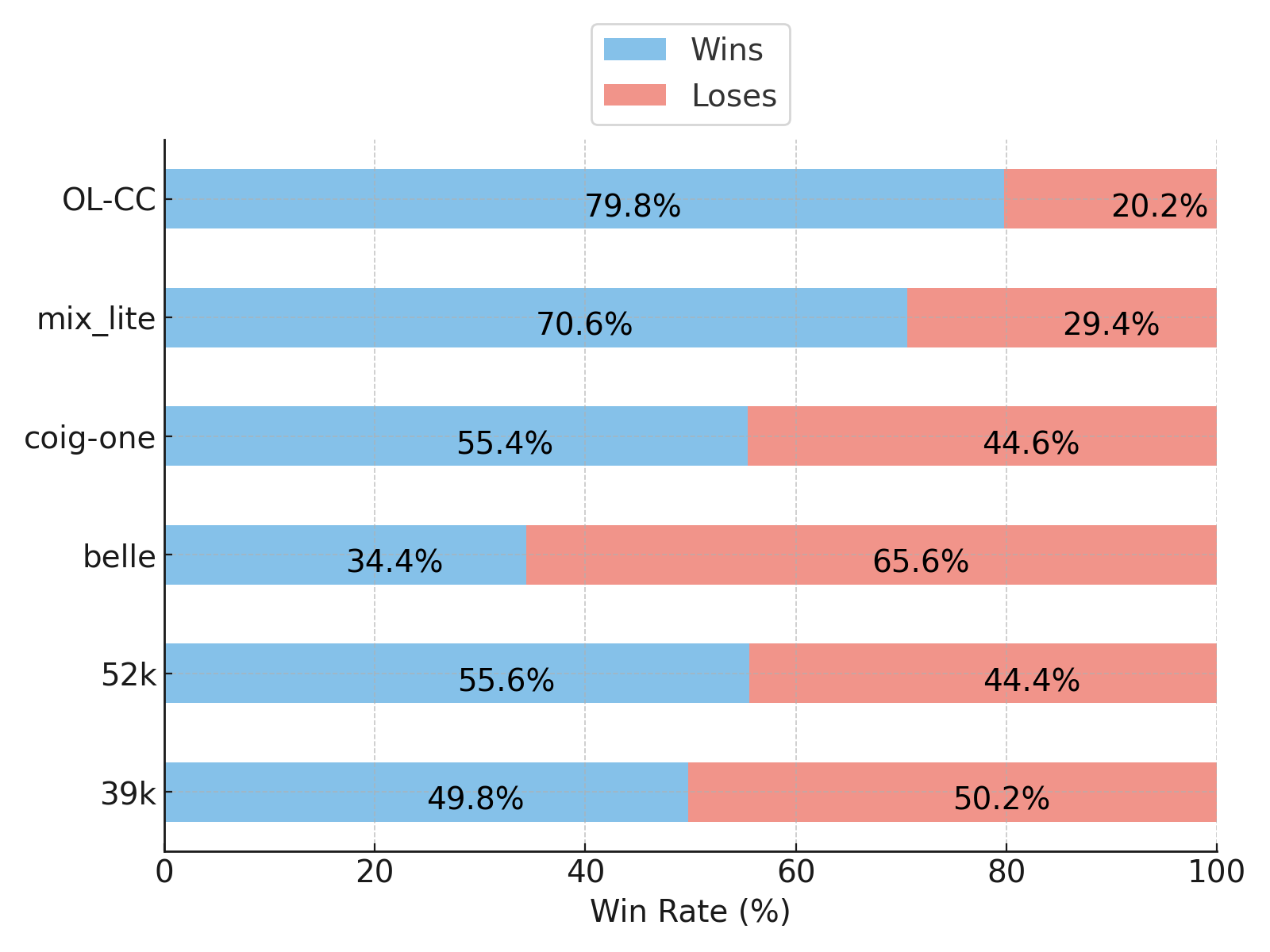}
        \label{fig:mix}
    \end{minipage}
    \caption{shows the human preference evaluation win rates of \emph{Kun-52k} and the \emph{mixed}(kun+COIG-one) model, respectively, when compared with other models. Notably, the \emph{mixed} model demonstrates a higher win rate than \emph{Kun-52k}, indicating its enhanced effectiven}
    \label{fig:mix_52k}
\end{figure*}

Further analysis is provided in Figures \ref{fig:mix_52k}, which depict the human preference evaluation win rates of \emph{Kun-52k} and the mixed model, respectively, when compared with other models. Notably, the mixed model demonstrates a higher win rate than \emph{Kun-52k}, indicating its enhanced effectiven.

\subsection{Prompt}
\begin{figure*}[ht]
  \centering
    \centering
    \includegraphics[width=1\linewidth]{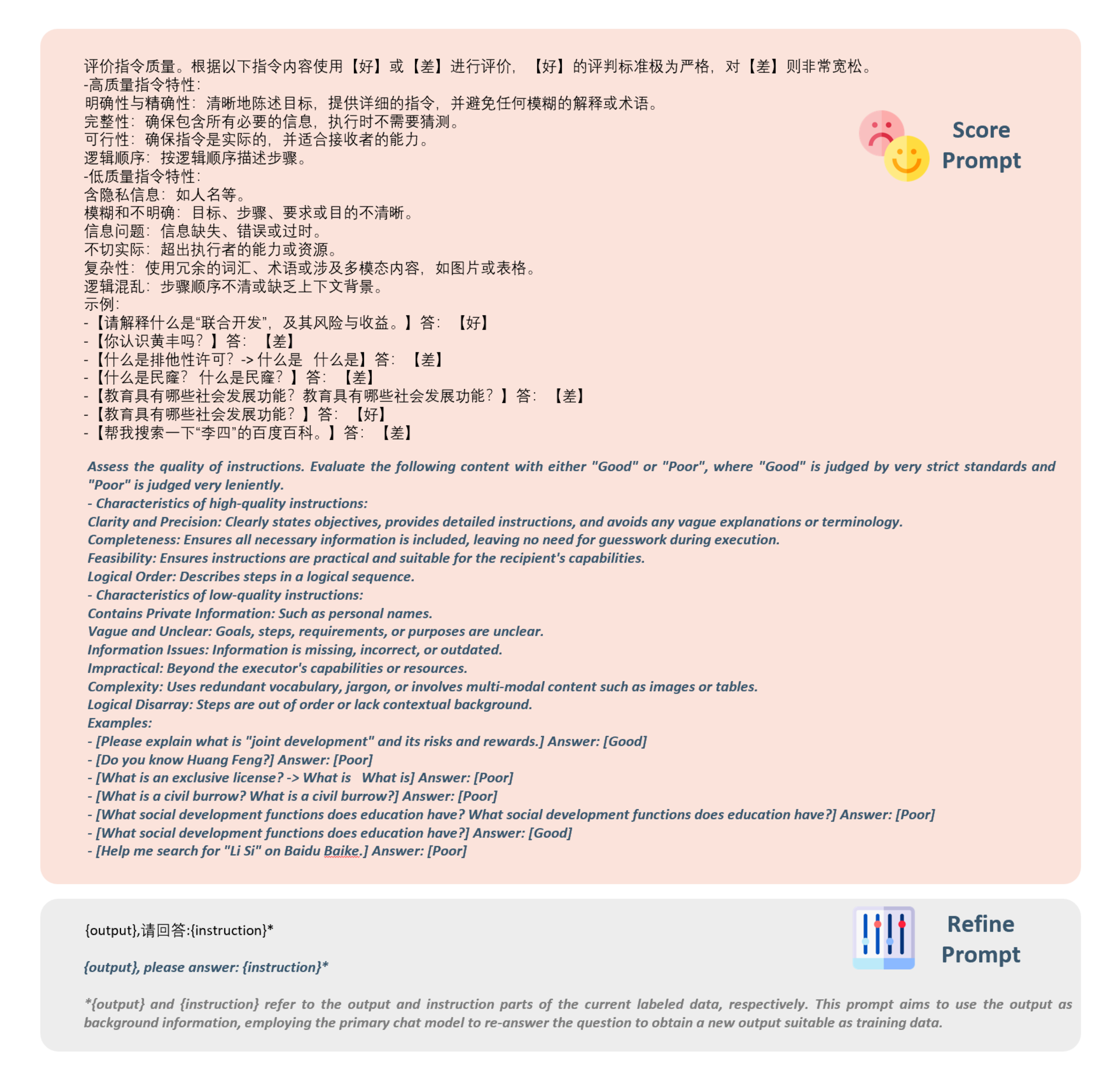}
  \hfill
    \centering
  \caption{The top section comprises score prompt used to assess the quality of labeled data instructions. The bottom section features refine prompt for refining the output part of the labeled data.}
  \label{appendix:score_and_refine}
\end{figure*}
\end{document}